\documentclass[letterpaper, 10 pt, conference]{ieeeconf}  

\IEEEoverridecommandlockouts                              

\usepackage{amsmath}
\usepackage{array}
\usepackage{afterpage}
\usepackage{booktabs}
\usepackage{float}
\usepackage{graphicx}
\usepackage{multirow}
\usepackage{placeins}
\usepackage{tabularx}
\usepackage{url}

\newcommand{\videoCount}{30}
\newcommand{\coveredPhysicalTask}{59}
\newcommand{\coveredTask}{91}

\newenvironment{subcategorylist}{%
    \begin{list}{$\bullet$}{%
        \setlength{\leftmargin}{1.1em}%
        \setlength{\labelwidth}{0.45em}%
        \setlength{\labelsep}{0.3em}%
        \setlength{\itemsep}{0pt}%
        \setlength{\parsep}{0pt}%
        \setlength{\topsep}{1pt}%
        \setlength{\partopsep}{0pt}%
    }%
}{\end{list}}

\title{\LARGE \bf
A Taxonomy of Construction Task Activities for Robot Workers
}

\author{Sadman Sakib$^{1}$, Zhangyi Peng$^{1, 3}$, Yujie Pang$^{1, 3}$, Yu Otsuki$^{2}$, and Mohammad Abdullah Al Faruque$^{1}$%
\thanks{$^{1,2}$Center for Resilient Autonomous Systems, University of California, Irvine (UCI), Irvine, CA, USA.}%
\thanks{$^{1}$Nhu Department of Electrical Engineering and Computer Science.\newline
\texttt{\{ssakib,znpeng,yujiep2,alfaruqu\}@uci.edu}}%
\thanks{$^{2}$Department of Civil and Environmental Engineering.\newline
\texttt{yu.otsuki@uci.edu}}%
\thanks{$^{3}$ The authors contributed during the UCInspire program.}%
}

\begin{document}

\maketitle
\thispagestyle{empty}
\pagestyle{empty}

\begin{abstract}
Recent vision--language--action models offer a path toward robots with broader
capabilities than conventional task-specific systems. Deploying such systems in
construction, however, requires a precise inventory of worker activities and the
capabilities needed to execute them. We present TARCAT, an occupation-grounded
taxonomy derived from \coveredTask{} O*NET task statements across seven
high-employment construction occupations and \videoCount{} instructional videos.
TARCAT defines 41 primitives spanning intellectual, social, and physical categories and provides a mechanism for composing parameterized primitive
sequences into reusable skills. This human-interpretable vocabulary supports the
specification of robot requirements and enables coding agents to retrieve and
extend skill libraries for task execution. We also demonstrate selected primitives on a DOBOT CR3 arm with a CRAFT hand. TARCAT thus provides a common
vocabulary for analyzing construction work and developing general-purpose
construction robots. Annotations are available at
\url{https://github.com/AICPS/TARCAT-Taxonomy}.
\end{abstract}

\section{INTRODUCTION}
Construction robotics has traditionally focused on purpose-built systems designed for specific tasks, such as ceiling installation, wall painting, and bricklaying~\cite{saidi2016robotics}. Meanwhile, current construction workflows are largely organized around human workers, who perform diverse activities involving different tools, actions, and interactions. As a result, general-purpose robotic systems capable of performing a broader range of construction activities could better assist human workers.

Recent advances in robotics, such as vision-language-action (VLA) models, are creating new opportunities for general-purpose robots capable of performing a broader range of tasks. Emerging agentic robotic systems, such as ASPIRE~\cite{lu2026aspire}, can also autonomously generate control programs, retain successful procedures in skill libraries, and retrieve them for new tasks. However, developing such systems still requires a systematic understanding of what tasks robots need to perform, what physical actions and skills those tasks require, and how these low-level capabilities can be composed into higher-level activities. 

Structured action taxonomies provide this interface between task activities and robot learning~\cite{pantano2022capability}. A manipulation vocabulary developed in \cite{paulius2019manipulation} includes 32 cooking motions characterized by contact, trajectory, duration, and manual-operation attributes. AVA~\cite{gu2018ava} annotates 80 atomic actions in movies. In industrial and manufacturing robotics, manipulation work has been classified into reusable tasks, skills, and actions~\cite{lee2024human,johannsmeier2025process}. TARMAC~\cite{huang2025tarmac} decomposes chemistry procedures into reusable robot manipulation actions. 

In the construction industry, taxonomy development for robotic tasks is still emerging. Everett and Slocum~\cite{everett1994automation} organized a set of 12 basic tasks that describe construction field work. Built on these basic tasks, Li and Leicht~\cite{li2026onet} recently analyzed O*NET~\cite{onet2026database} task statements across 15 construction occupations to identify task verbs, objectives, tools, and materials, but the analysis does not consider physical robot embodiments. At the human-motion level, Rodrigues et al.~\cite{rodrigues2023multidimensional} classified human--robot interaction according to platform, autonomy, interaction, and team structure, while the study does not connect occupation-grounded construction tasks to robotic embodiments and reusable robot actions.

\begin{figure*}[t]
    \centering
    \includegraphics[width=\textwidth]{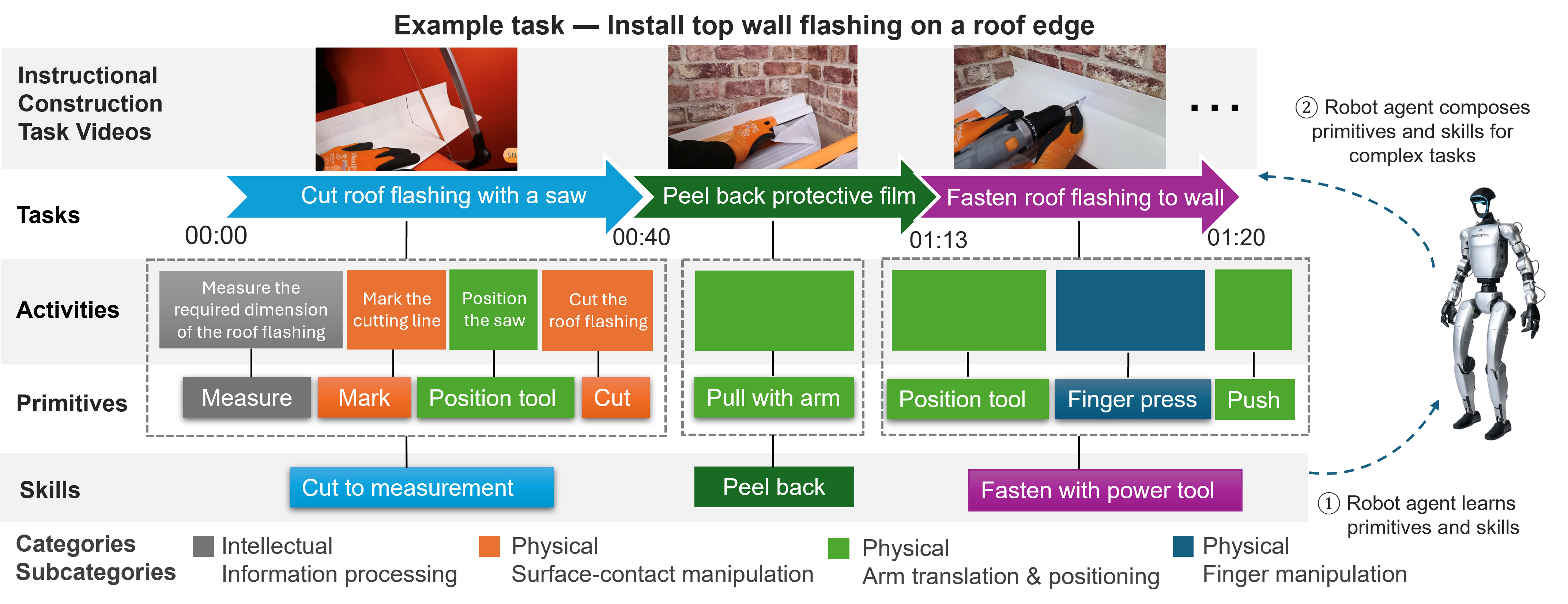}
    \caption{\textbf{Overview of TARCAT.}
    Instructional construction videos are segmented into observable
    activities, labeled with reusable primitives, and organized into
    skills; activity and primitive colors indicate the category and subcategory in Table~\ref{tab:action-primitives}, and each skill's color matches the task segment it accomplishes. Using the developed taxonomy and annotations, robot agents can (i) learn primitives and skills and
    (ii) compose them to execute complex tasks.}
    \label{fig:tarcat-overview}
\end{figure*}

We propose TARCAT (Taxonomy of Construction Task Activities for Robot Workers)
to address these gaps. TARCAT provides an occupation-grounded vocabulary of construction tasks, their composition into reusable robotic actions, and video annotations (Fig.~\ref{fig:tarcat-overview}).
TARCAT contributes (1) 41 primitives spanning intellectual, social, and physical categories;
(2) annotations of \videoCount{} videos and labels covering \coveredTask{} O*NET tasks across seven high-employment construction occupations; and (3) hardware demonstrations on a collaborative robotic arm with a dexterous hand.

\newcommand{\actionprimitivestable}{%
\begin{table*}[!t]
\caption{TARCAT primitives organized by category and subcategory, with example activities.}
\label{tab:action-primitives}
\centering
\fontsize{8.5}{9.5}\selectfont
\setlength{\tabcolsep}{0pt}
\setlength{\aboverulesep}{0.2ex}
\setlength{\belowrulesep}{0.2ex}
\renewcommand{\arraystretch}{1.02}
\begin{tabularx}{\textwidth}{@{}
    >{\raggedright\arraybackslash}p{0.11\textwidth}
    @{\hspace{8pt}}
    >{\raggedright\arraybackslash}p{0.13\textwidth}
    @{\hspace{8pt}}
    >{\raggedright\arraybackslash}p{0.34\textwidth}
    @{\hspace{8pt}}
    >{\raggedright\arraybackslash}X@{}}
\toprule
\textbf{Category} & \textbf{Subcategory} & \textbf{Primitives} & \textbf{Example Activities} \\
\midrule
\multirow{9}{0.11\textwidth}{Intellectual}
& \multirow{4}{0.13\textwidth}{Information processing}
& Take measurements
& Read pressure gauge; measure roof width \\
& & Study plan, instruction or specifications
& Read compressor label; check breaker rating \\
& & Listen to or observe instructions or signals
& Follow spoken directions; observe crane signals \\
& & Prepare records, documents, and reports
& Write project report; record inspection result \\
\cmidrule{2-4}
& \multirow{5}{0.13\textwidth}{Problem solving}
& Determine tool-application location
& Locate drill point; locate nail-gun cap \\
& & Inspect to check state or find issues
& Inspect roof damage; check ladder locks \\
& & Estimate time, materials, labor and cost
& Estimate roofing labor; calculate mortar quantity \\
& & Select and procure tools and materials
& Order lumber; choose paint finish \\
& & Prepare design, layout or installation plans
& Plan wiring layout; sketch equipment locations \\
\midrule
\multirow{3}{0.11\textwidth}{Social}
& \multirow{3}{0.13\textwidth}{Coordinating}
& Talk to convey information
& Discuss work plan; explain safety issue \\
& & Use gestures to signal other workers
& Signal crane operator; direct team lift \\
& & Schedule and coordinate work with others
& Schedule work crew; coordinate subcontractors \\
\midrule
\multirow{29}{0.11\textwidth}{Physical}
& \multirow{4}{0.13\textwidth}{Finger manipulation}
& Pull or push an object with fingers
& Open nail-gun cap; connect air hose \\
& & Press an object with fingers
& Press laser button; squeeze nail-gun trigger \\
& & Open fingers
& Release plumb bob; release stud finder \\
& & Rotate an object with fingertips
& Adjust pressure valve; tighten spray-can straw \\
\cmidrule{2-4}
& \multirow{5}{0.13\textwidth}{Arm translation and positioning}
& Grab an object
& Catch plumb-bob string; grab panel cover \\
& & Pull or push an object with arm
& Open cabinet door; push breaker into slot \\
& & Pick or place an object
& Pick up trowel; place electrical box \\
& & Position an object
& Align roof flashing; position screwdriver \\
& & Move tool along a linear path
& Apply joint sealant; dip brush into paint \\
\cmidrule{2-4}
& \multirow{6}{0.13\textwidth}{Arm and wrist rotation}
& Twist an object using wrist rotation
& Open cement bottle; twist wire with pliers \\
& & Pivot a tool about a fixed point
& Tighten bolt with wrench; turn valve key \\
& & Swing arm while holding an object
& Hammer a nail; chip mortar with chisel \\
& & Reorient an object
& Flip shingle; turn brick over \\
& & Pour liquid from a container
& Pour paint into tray; pour water into bucket \\
& & Scoop material with a tool
& Scoop mortar with trowel; scoop soil with shovel \\
\cmidrule{2-4}
& \multirow{3}{0.13\textwidth}{Mixing motion}
& Shake an object
& Shake spray can; shake mixing container \\
& & Stir liquid and solid mixture
& Mix mortar with trowel; stir paint with stick \\
& & Wrap around, tie or untie
& Tie rope around ladder; untie plumb-bob knot \\
\cmidrule{2-4}
& \multirow{4}{0.13\textwidth}{Surface-contact manipulation}
& Mark on a surface
& Mark stud line; mark screw point \\
& & Cut an object
& Saw wooden stud; slice roofing shingle \\
& & Spread object on surface
& Spread mortar; smooth joint compound \\
& & Stroke tool on surface
& Brush paint on wall; roll paint on ceiling \\
\cmidrule{2-4}
& \multirow{3}{0.13\textwidth}{Whole-body locomotion}
& Walk to a location
& Walk to inspection area; approach electric outlet \\
& & Carry an object to a location
& Carry measuring rod; carry roll of felt \\
& & Climb up or down
& Climb ladder; descend scaffold \\
\cmidrule{2-4}
& \multirow{4}{0.13\textwidth}{Whole-body load handling}
& Pull or push a heavy object
& Align wall cabinet; slide loaded pallet \\
& & Pick or place a heavy object
& Pick up ladder; place wall cabinet \\
& & Carry a heavy object
& Carry large pipe; transport ladder \\
& & Bend down or straighten up
& Pick up leaf blower from ground; lift drill and stand \\

\bottomrule
\end{tabularx}
\end{table*}
}
\actionprimitivestable

\section{OCCUPATIONAL DATA AND INSTRUCTIONAL VIDEO COLLECTION}
Our taxonomy framework is grounded in two U.S. Department of Labor databases. A recent study, GDPval~\cite{openai2025gdpval}, used BLS employment data and O*NET task statements to identify economically significant knowledge-work tasks. However, GDPval did not consider construction occupations or tasks involving physical movement. TARCAT adopts a similar grounding strategy to identify economically significant construction occupations rather than relying on a manually selected set of construction tasks.

The BLS Occupational Outlook Handbook~\cite{bls2025construction} reports 2024 employment data for Construction and Extraction occupations. Based on the BLS data, we selected the seven construction occupations with the largest number of employed workers. O*NET~\cite{onet2026database} is another database that provides SOC-aligned task statements for each occupation. Using the BLS-to-SOC occupation mapping provided by the BLS~\cite{bls2025construction}, we selected the most relevant O*NET SOC occupation for each of the seven BLS construction occupations. We identified 169 O*NET task statements corresponding to these seven O*NET SOC occupations: (i) Construction Laborers, (ii) Carpenters, (iii) Electricians, (iv) Plumbers, Pipefitters, and Steamfitters, (v) Painters, Construction and Maintenance, (vi) Brickmasons and Blockmasons, and (vii) Roofers. Note that the heavy machinery operator occupation was excluded because it falls outside the scope of this study.

We classified 137 of the 169 O*NET task statements as involving physical movement and the remaining 32 as non-movement tasks. Using ChatGPT with GPT-5.6~\cite{openai2026gpt56}, we searched YouTube for instructional demonstrations of the movement tasks. We manually retained \videoCount{} videos with observable, task-relevant actions and mapped each video to every task statement it demonstrated. These videos cover \coveredPhysicalTask{} movement tasks. We annotated all 32 non-movement tasks directly from their task statements, yielding coverage of \coveredTask{} O*NET tasks across the seven occupations.

\section{TAXONOMY OF CONSTRUCTION TASKS}
\subsection{Task and Activity}
This section clarifies the terminology used in our taxonomy.
First, O*NET defines task statements as ``\textit{task statements are exemplars of the types of activities workers in an occupation may perform and are typically conceptualized as the smallest unit of activity with a meaningful outcome}''\cite{onet2026database}. Hence, O*NET uses \textit{activity} to broadly describe work performed by a human worker, not as a level below a task.

While O*NET task statements are adopted in this study, we use the term \textit{activity} in a more specific sense to denote a finer-grained unit of task execution. Accordingly, we define the relationship between a \textit{task} and an \textit{activity} as follows: A task is an externally specified unit comprising an activity or a sequence of activities to achieve a goal. This distinction in the definition of activity is necessary because O*NET task statements, while sufficiently detailed for human occupational analysis, may remain too coarse for robot execution and therefore require further decomposition. Note that a random sequence of activities without an intended goal does not constitute a task.

We further introduce a nested task structure, in which a task can be decomposed into smaller tasks, each with its own goal. The level of decomposition depends on the agent's capabilities. For example, O*NET task statements suffice for human workers, whereas robot agents may require smaller, short-horizon tasks compatible with their capabilities.

As an example, consider the task shown in Fig.~\ref{fig:tarcat-overview}, ``Install top wall flashing on a roof edge'', which is decomposed into three smaller tasks. The first task, ``Cut roof flashing with a saw'' shown in Fig.~\ref{fig:tarcat-overview}, can be decomposed into four activities: (i) measure the required dimension of the roof flashing, (ii) mark the cutting line, (iii) position the saw, and (iv) cut the roof flashing. Together, these activities achieve the goal of the first task: the roof flashing is cut to the required dimension. 

\subsection{Category and Subcategory}
We classify the activities associated with the \coveredTask{} O*NET tasks into three categories: (1) Intellectual, (2) Social, and (3) Physical, following the worker-task taxonomy proposed by~\cite{fernandezmacias2022taxonomy}. The taxonomy~\cite{fernandezmacias2022taxonomy} further divides each category into subcategories. We retain all subcategories defined for (1) Intellectual and adopt the Coordinating subcategory from (2) Social as relevant to construction tasks. The definitions of these categories and subcategories follow~\cite{fernandezmacias2022taxonomy}.

For the (3) Physical category, the study~\cite{fernandezmacias2022taxonomy} defines three subcategories: strength, dexterity, and navigation based on the physical capability required of workers. To expose finer robot-action requirements, TARCAT instead distinguishes the body segment primarily involved and the form of motion or contact, thereby defining a new set of fine-grained subcategories under the Physical category as follows:
\begin{subcategorylist}
    \item \textbf{Finger manipulation:} Manipulate primarily through
    finger motion or force, with little gross arm movement.

    \item \textbf{Arm translation and positioning:} Grab or move an object
    through coordinated hand and arm motion.

    \item \textbf{Arm and wrist rotation:} Rotate or reorient the arm,
    wrist, tool, or object to apply force.

    \item \textbf{Mixing motion:} Use repeated, circular, or wrapping
    motions to mix or secure materials.

    \item \textbf{Surface-contact manipulation:} Move a tool while
    maintaining controlled contact with a work surface.

    \item \textbf{Whole-body locomotion:} Use the legs and torso to move
    between locations or transport an object.

    \item \textbf{Whole-body load handling:} Use whole-body force and
    postural control to manipulate heavy objects or bend and straighten.
\end{subcategorylist}

Table I summarizes the classification of example activities derived from the O*NET task statements into three categories and their corresponding subcategories.

\subsection{Activity, Primitive, and Parameter}
This section introduces the terms \textit{primitive} and \textit{parameter} and relates them to \textit{activity}. A \emph{primitive} is the smallest atomic unit, such as a physical action, perception, or cognitive process, that performs a distinct operation and can be reused across contexts. \emph{Parameters} configure a primitive for a specific activity or task and may specify objects, tools, targets, or other context-dependent information. We then define an \textit{activity} as an instantiation of a primitive with context-specific parameters. For example, the activity ``Read pressure gauge'' can be represented by the primitive ``Take measurements'' with ``pressure gauge'' bound to the \texttt{to\_measure} parameter.

\begin{figure}[t]
    \centering
    \includegraphics[width=\columnwidth]{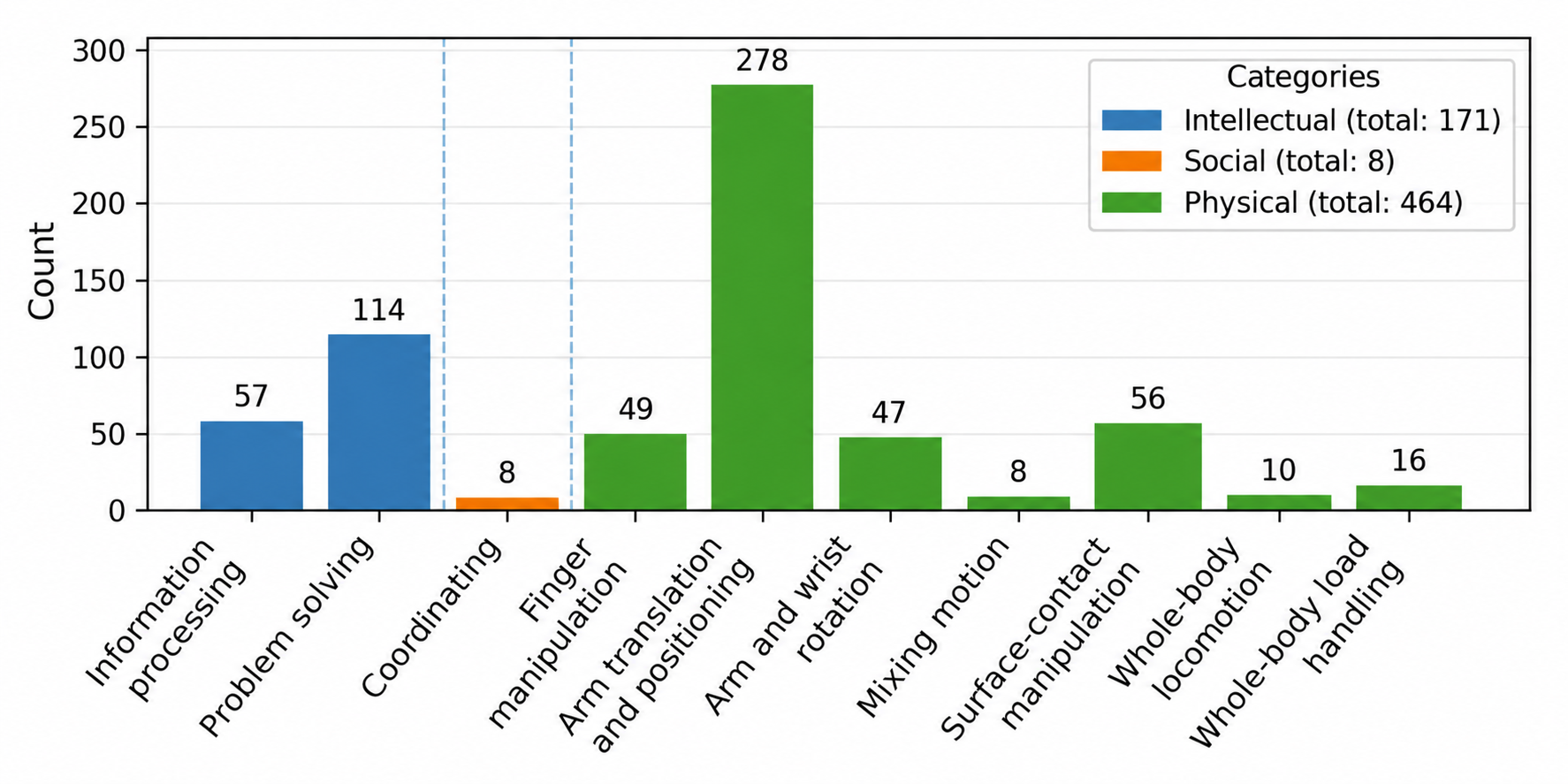}
    \caption{\textbf{Distribution of primitive occurrences across labeled
    O*NET tasks and instructional videos.} Bars denote sub-categories, and
    colors denote their parent categories.}
    \label{fig:action-primitive-counts}
\end{figure}

Table~\ref{tab:action-primitives} lists the 41 unique primitives identified from activities in the covered O*NET task statements and instructional videos. A human annotator labeled time-stamped video segments with the corresponding primitives. Figure~\ref{fig:action-primitive-counts} summarizes the resulting distribution of 643 primitive labels from the annotated videos across the subcategories. ``Arm translation and positioning'' is the most frequent subcategory (278 occurrences), followed by ``Problem solving'' (114).
\subsection{Skill}
For the application of this taxonomy, we then define a \textit{skill} as follows: A skill is a unit of agent capability, acquired through learning or programming, to execute an activity or a sequence of activities to achieve a goal. While the definitions of task and skill are similar in that both involve achieving a goal through an activity or sequence of activities, the key distinction is that a task is externally specified to an agent, whereas a skill represents the agent's own capability. Therefore, a skill solves a task. In a scheduling framework, an agent can be selected for a given task based on whether it possesses the skill required to perform that task.

As an example of a skill, consider a robotic agent that has mastered the skill ``Fasten an object with a power tool'', which comprises four activities with associated primitives and parameters: (1) Position an object (\texttt{to\_position} = fixing hardware), (2) Position an object (\texttt{to\_position} = power tool), (3) Press an object with fingers (\texttt{to\_press} = power-tool trigger), and (4) Pull or push an object with arm (\texttt{to\_pull\_push} = power tool). 

Similar to the nested task structure, skills can also be recursively composed, allowing lower-level skills to be combined into higher-level skills.

\FloatBarrier

\section{PHYSICAL DEMONSTRATION}
\begin{figure}[t]
    \centering
    \includegraphics[width=\columnwidth]{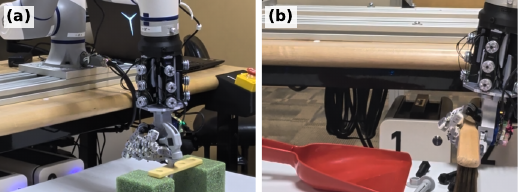}
    \caption{\textbf{Physical demonstrations of TARCAT primitives.}
    (a) Hammering a nail with a primitive ``\emph{Swing arm while holding an object}''.
    (b) Brushing objects toward a dustpan with a primitive ``\emph{Stroke tool on surface}''.}
    \label{fig:physical-demonstrations}
\end{figure}

To assess executability, we selected four activities from the Physical category: positioning a
power screwdriver, placing a hammer in a box, hammering a nail, and brushing
objects aside. We mapped each activity to its TARCAT primitive and
implemented scripted demonstrations on a six-axis DOBOT CR3
arm~\cite{dobot2026cr3} with a 15-actuator, tendon-driven CRAFT
hand~\cite{lin2026craft}. Figure~\ref{fig:physical-demonstrations} shows two
representative executions. Collectively, the demonstrations cover arm
translation and positioning, arm and wrist rotation, and surface-contact
manipulation, providing initial evidence that these taxonomy elements can be
instantiated as executable physical behaviors. The demonstration video is
available at \url{https://youtu.be/Dc-bcWGwTLc}.

Future work will expand the corpus to evaluate long-horizon skills and develop data-centric methods for classifying primitives~\cite{liconti2026benchmark}.

\bibliographystyle{IEEEtran}
\bibliography{references}

@article{huang2025tarmac,
  author        = {Kefeng Huang and Jonathon Pipe and Alice E. Martin and Tianyuan Wang and Barnabas A. Franklin and Andy M. Tyrrell and Ian J. S. Fairlamb and Jihong Zhu},
  title         = {{TARMAC}: A Taxonomy for Robot Manipulation in Chemistry},
  journal       = {arXiv preprint arXiv:2510.19289},
  year          = {2025},
  doi           = {10.48550/arXiv.2510.19289}
}

@article{everett1994automation,
  author        = {John G. Everett and Alexander H. Slocum},
  title         = {Automation and Robotics Opportunities: Construction versus Manufacturing},
  journal       = {Journal of Construction Engineering and Management},
  volume        = {120},
  number        = {2},
  pages         = {443--452},
  year          = {1994},
  doi           = {10.1061/(ASCE)0733-9364(1994)120:2(443)}
}

@article{rodrigues2023multidimensional,
  author        = {Patrick Rodrigues and Rashmi Singh and Mert Oytun and Pooya Adami and Peter Woods and Burcin Becerik-Gerber and Lucio Soibelman and Yasemin Copur-Gencturk and Gale Lucas},
  title         = {A Multidimensional Taxonomy for Human--Robot Interaction in Construction},
  journal       = {Automation in Construction},
  volume        = {150},
  pages         = {104845},
  year          = {2023},
  doi           = {10.1016/j.autcon.2023.104845}
}

@techreport{openai2025gdpval,
  author        = {{OpenAI}},
  title         = {{GDPval}: Evaluating {AI} Model Performance on Real-World Tasks},
  institution   = {OpenAI},
  year          = {2025},
  url           = {https://openai.com/index/gdpval/}
}

@misc{openai2026gpt56,
  author        = {{OpenAI}},
  title         = {{GPT-5.6}: Frontier Intelligence that Scales with Your Ambition},
  year          = {2026},
  month         = jul,
  url           = {https://openai.com/index/gpt-5-6/},
  note          = {Accessed: 2026-08-23}
}

@misc{bls2025construction,
  author        = {{U.S. Bureau of Labor Statistics}},
  title         = {Construction and Extraction Occupations},
  howpublished  = {Occupational Outlook Handbook},
  year          = {2025},
  url           = {https://www.bls.gov/ooh/construction-and-extraction/home.htm},
  note          = {Accessed: 2026-08-20}
}

@misc{onet2026database,
  author        = {{National Center for O*NET Development}},
  title         = {{O*NET} Database},
  year          = {2026},
  url           = {https://www.onetcenter.org/database.html},
  note          = {Accessed: 2026-08-20}
}

@article{lin2026craft,
  author        = {Leo Lin and Shivansh Patel and Jay Moon and Svetlana Lazebnik and Unnat Jain},
  title         = {{CRAFT}: A Tendon-Driven Hand with Hybrid Hard--Soft Compliance},
  journal       = {arXiv preprint arXiv:2603.12120},
  year          = {2026},
  doi           = {10.48550/arXiv.2603.12120}
}

@misc{dobot2026cr3,
  author        = {{Dobot}},
  title         = {{DOBOT CR3} Specifications},
  year          = {2026},
  url           = {https://www.dobot.us/cr3-specs/},
  note          = {Accessed: 2026-08-20}
}

@inproceedings{paulius2019manipulation,
  author        = {David Paulius and Yongqiang Huang and Jason Meloncon and Yu Sun},
  title         = {Manipulation Motion Taxonomy and Coding for Robots},
  booktitle     = {2019 IEEE/RSJ International Conference on Intelligent Robots and Systems (IROS)},
  pages         = {5596--5601},
  year          = {2019},
  doi           = {10.1109/IROS40897.2019.8967754}
}

@inproceedings{gu2018ava,
  author        = {Chunhui Gu and Chen Sun and David A. Ross and Carl Vondrick and Caroline Pantofaru and Yeqing Li and Sudheendra Vijayanarasimhan and George Toderici and Susanna Ricco and Rahul Sukthankar and Cordelia Schmid and Jitendra Malik},
  title         = {{AVA}: A Video Dataset of Spatio-Temporally Localized Atomic Visual Actions},
  booktitle     = {2018 IEEE/CVF Conference on Computer Vision and Pattern Recognition (CVPR)},
  pages         = {6047--6056},
  year          = {2018},
  doi           = {10.1109/CVPR.2018.00633}
}

@article{lu2026aspire,
  title={ASPIRE: Agentic/Skills Discovery for Robotics},
  author={Lu, Runyu and Wu, Yubo and Kou, Ethan and Fu, Letian and Xiao, Wenli and Mandlekar, Ajay and Xu, Yinzhen and Shi, Guanya and Goldberg, Ken and Chen, Ang and others},
  journal={arXiv preprint arXiv:2607.00272},
  year={2026}
}

@incollection{saidi2016robotics,
  title={Robotics in construction},
  author={Saidi, Kamel S and Bock, Thomas and Georgoulas, Christos},
  booktitle={Springer handbook of robotics},
  pages={1493--1520},
  year={2016},
  publisher={Springer}
}

@article{lee2024human,
  title={Human-robot shared assembly taxonomy: A step toward seamless human-robot knowledge transfer},
  author={Lee, Regina Kyung-Jin and Zheng, Hao and Lu, Yuqian},
  journal={Robotics and Computer-Integrated Manufacturing},
  volume={86},
  pages={102686},
  year={2024},
  publisher={Elsevier}
}

@article{johannsmeier2025process,
  title={A process-centric manipulation taxonomy for the organization, classification and synthesis of tactile robot skills},
  author={Johannsmeier, Lars and Schneider, Samuel and Li, Yanan and Burdet, Etienne and Haddadin, Sami},
  journal={Nature Machine Intelligence},
  volume={7},
  number={6},
  pages={916--927},
  year={2025},
  publisher={Nature Publishing Group UK London}
}

@article{fernandezmacias2022taxonomy,
  author        = {Enrique Fern{\'a}ndez-Mac{\'i}as and Martina Bisello},
  title         = {A Comprehensive Taxonomy of Tasks for Assessing the Impact of New Technologies on Work},
  journal       = {Social Indicators Research},
  volume        = {159},
  number        = {2},
  pages         = {821--841},
  year          = {2022},
  publisher     = {Springer},
  doi           = {10.1007/s11205-021-02768-7}
}

@inproceedings{li2026onet,
  author        = {Fangxiao Li and Robert M. Leicht},
  title         = {Exploring Construction Tasks That Can Be Performed by Robots},
  booktitle     = {Construction Research Congress 2026},
  pages         = {120--129},
  year          = {2026},
  month         = aug,
  publisher     = {American Society of Civil Engineers},
  doi           = {10.1061/9780784486979.012}
}

@article{liconti2026benchmark,
  title={A benchmark of dexterity for anthropomorphic robotic hands},
  author={Liconti, Davide and Zhou, Yuning and Toshimitsu, Yasunori and Hinchet, Ronan and Katzschmann, Robert K},
  journal={arXiv preprint arXiv:2604.09294},
  year={2026}
}

@inproceedings{pantano2022capability,
  title={Capability-based frameworks for industrial robot skills: a survey},
  author={Pantano, Matteo and Eiband, Thomas and Lee, Dongheui},
  booktitle={2022 IEEE 18th International Conference on Automation Science and Engineering (CASE)},
  pages={2355--2362},
  year={2022},
  organization={IEEE}
}



\end{document}